# Modeling Effect of Lockdowns and Other Effects on India Covid-19 Infections Using SEIR Model and Machine Learning

Sathiyanarayanan Sampath
Global AI Accelerator
Ericsson
Bangalore, India
sathiyanarayanan@ericsson.com

Joy Bose
Global AI Accelerator
Ericsson
Bangalore, India
joy.bose@ericsson.com

*Abstract*— The SEIR model is a widely used epidemiological model used to predict the rise in infections. This model has been widely used in different countries to predict the number of Covid-19 cases. But the original SEIR model does not take into account the effect of factors such as lockdowns, vaccines, and re-infections. In India the first wave of Covid started in March 2020 and the second wave in April 2021. In this paper, we modify the SEIR model equations to model the effect of lockdowns and other influencers, and fit the model on data of the daily Covid-19 infections in India using lmfit, a python library for least squares minimization for curve fitting. We modify R0 parameter in the standard SEIR model as a rectangle in order to account for the effect of lockdowns. Our modified SEIR model accurately fits the available data of infections.

*Keywords—Covid-19, SEIR model, lockdowns, machine learning, least squares minimization*

## I. INTRODUCTION

The global Covid-19 pandemic has caused more than 32 million infections and taken the lives of 0.4 million people in India alone so far, as per reports [1]. In India the first wave of Covid-19 in 2020 and the second wave in 2021 caused loss of lives, which for future waves can be alleviated if we could accurately model the number of infections in advance and take measures to control them.

In the past century alone, the world has been witness to other deadly pandemics such as the Spanish Flu, SARS and MERS. Now thanks to the advance of data science and epidemiological models [2] like SIR and SEIR [2, 3], we can have the ability to model the spread of such pandemics more rapidly and accurately.

In this paper, we use a modified form of the SEIR model to model the first and second wave of Covid-19 infections in India and also take into account the effects of factors that influence the number of infected persons such as lockdown, vaccines and second infections. We then fit the modified model to the available data of infections in India to ascertain how well can the model fit the data.

The remaining sections of the paper are as follows: section 2 contains related work in the area of SEIR and Covid-19 modeling. Section 3 describes our changes to the SEIR model to account for lockdowns and other factors. Section 4 gives the datasets and parameters used in our model. Section 5 gives the result of our curve fitting using the modified SEIR model on Covid-19 data. Section 6 concludes the paper and gives some ideas about further work planned.

## II. BACKGROUND AND RELATED WORK

In this section, we consider some background work related to SEIR model.

### A. SEIR model: Original SEIR equations

The SEIR model is an epidemiological model. It was proposed by Aron and I.B. Schwartz [3, 4]. It models the population that is affected by epidemics and pandemics like Covid by breaking into 4 parts: S (population that is susceptible), E (population that is exposed to the disease), I (population that is infected by the disease), and R (population that is recovered from the disease). In the original SEIR model, all four terms S, E, I and R are fractions and add to 1, but in our case, similar to [4] we have taken them as absolute numbers that add up to N, size of the population.

Figure 1 shows a block diagram of the original SEIR model and its fractions.

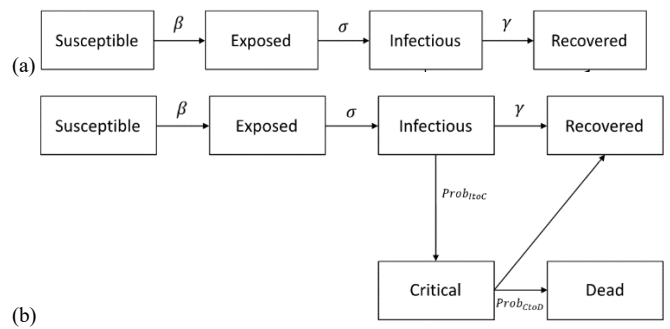

Fig. 1. Block diagram for the (a) Original SEIR model and (b) the SEIR mdoel that takes critical cases and deaths into account (taken from [4])

The SEIR model has differential equations (shown in equations 1-6 below) to model the percentage of the population that falls under the above fractions and how they change with time. $\beta, \sigma$ and $\gamma$ are parameters that vary with time. This is taken from the blog post by Froese [4]:

Notations

N Total population of a country or region under consideration

C Number of people critically ill, as a function of time

D Number of people dying in a day, as a function of time

$\beta$ – It is the transmission rate, the number of people on average to which each individual transmits the virus per unit time. Time is generally measured in units of days. $\beta$ is a function of $R_0$

$R_0$ is the number of people whom an infected person can infect in one time unit (i.e. per day).

$\sigma$ Incubation period

$\gamma$ Recovery rate – Proportion of infected individuals who are recovering per unit time (i.e. per day)

$Prob_{ItoC}$ Probability of infected becoming critical

$Prob_{CtoD}$ Probability of critical people dying

$Beds$ No of ICU beds available to treat critical Covid patients, it is a function of time t

$\frac{dR}{dt}$ Population gaining permanent immunity

$$\frac{dS}{dt} = \frac{-\beta IS}{N} \quad (1)$$

$$\frac{dE}{dt} = \frac{\beta IS}{N} - \sigma E \quad (2)$$

$$\frac{dI}{dt} = \sigma E - 1/12.0 * Prob_{ItoC} * I - \gamma(1 - Prob_{ItoC}) * I \quad (3)$$

$$\frac{dC}{dt} = 1/12.0 * Prob_{ItoC} * I - 1/7.5 * Prob_{CtoD} * \min(Beds, C) - \max(0, C - Beds) - (1 - Prob_{CtoD}) \, 1/6.5 * \min(Beds - C) \quad (4)$$

$$\beta = \gamma * R_0 \quad (5)$$

$$\frac{dR}{dt} = \gamma(1 - Prob_{ItoC}) * I + (1 - Prob_{CtoD}) \, 1/6.5 * \min(Beds - C) \quad (6)$$

$$\frac{dD}{dt} = 1/7.5 * Prob_{CtoD} * \min(Beds, C) + \max(0, C - Beds) \quad (7)$$

R0 is the number of people infected by an already infected person in a unit time i.e. one day. It is a function of time. It is given in (8):

$$R0 = \frac{R0_{start} - R0_{end}}{(1 + e^{-k*(-t+t0)})} + R0_{end} \quad (8)$$

Here the equation (8) represents a smooth continuous step function. Parameters $R0_{start}$ and $R0_{end}$ control the maximum and minimum values of the function, k influences the smoothness of the function, higher values of the k make the curve steeper, t is the time, t0 is the point of inflection, where the function takes average value between $R0_{start}$ and $R0_{end}$.

*B. Related work on modeling of Covid-19*

There have been a few studies in the literature that attempt to model the spread of Covid-19, both globally and in individual countries. In this subsection we discuss a few of these related works.

A number of studies, such as in [11-14] use standard epidemiological models such as SIR and SEIR to model the Covid data for different countries.

Bjørnstad et al [15] developed a time-series Susceptible Infected Recovered (TSIR) model, which includes stochasticity inherent in disease transmission and random immigration. The model fits well for data from 60 cities in England and wales. They found R0 (one of the parameters of the SEIR model) to be invariant across the cities of varying size.

The paper by Li and Muldowney [16] studies the non-linear incidence rates (rate of new infection) in epidemiology.

In the work by He et al [17], the SEIR model is applied to COVID-19 with control strategies like hospital, quarantine, etc. Particle swarm optimization (PSO) is used to find the parameters of the model and found it varies based on scenario. The authors propose an improved SEIR model with additional states like quarantine and hospitalization, which are control measures introduced by the government.

The work by Pandey et al [18] analyzes India's COVID-19 data and make predictions on a number of cases for future two weeks, using the standard SEIR model along with regression.

*C. Related work on modeling of effect of lockdowns*

There are also a few works about modeling the effects of lockdowns on Covid-19 data.

Qian et. al [8] used a Bayesian model using a 2-layer Gaussian process to predict the effect of lockdowns, using compartmentalized priors given in by standard SEIR model.

Bhardwaj et al [9] propose a reinforcement learning (RL) based approach by modeling the effect of lockdowns as a Markov Decision Process and using the discrete time SEIRD as the epidemiological model, where RL gives the optimal lockdown policy.

Bagal et al [10] estimate the parameters of the SIR model for Covid-19 data for India, especially to measure how much is the impact of lockdowns.

The above approaches do not, however, modify the SEIR equations and do not incorporate additional factors such as effect of vaccines and second infections, as we have done in this paper.

III. OUR APPROACH

In this section we provide details of our approach to modeling Covid infections.

For India-19 Covid infections, our data source is the John Hopkins University dataset [5] that is updated every day and carries information about number of infected, recovered etc in different countries. We have filtered out the data for India only.

The graph of the Covid-19 mortalities for India, taken from [5] is shown in fig. 2.

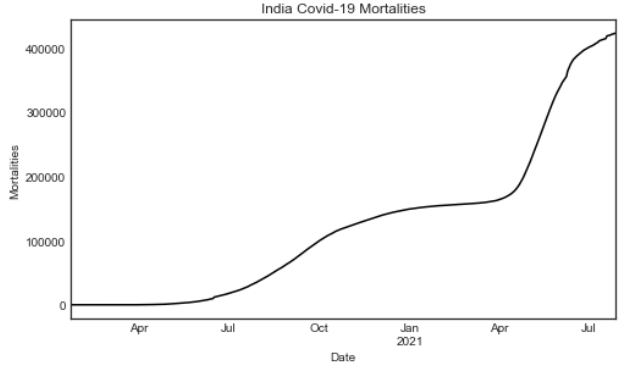

Fig. 2. Graph of the Covid-19 mortalities (deaths) in India in 2020 and 2021, taken from [5]

We modify the SEIR model to incorporate the effect of lockdowns and other factors. After this, we fit the modified equations to real world data of Covid-19 infections using a suitable fitting function.

A block diagram of the modified SEIR model is shown in fig. 3.

In the next few subsections we explain how we modify the SEIR equations for each of the different effects: lockdown, vaccination and second infections.

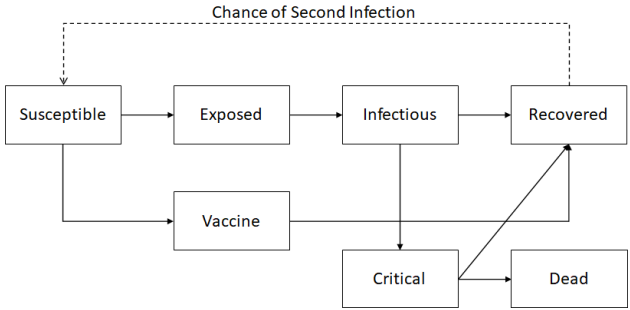

Fig. 3. Block diagram for the modified SEIR model that can model second infections and vaccines

### A. SEIR model: Modified for effect of lockdowns

To the original SEIR equation, we have added a term to represent the rectangle function. The shape of a smooth continuous rectangle function varying with time t is shown in fig. 4.

We have modified the SEIR equations to have R0 as a function of time to depict real world scenario. We have defined R0 with smooth rectangular function, which is continuously differentiable, which stays low when there is no spread, gradually increases when spread is high and stays high for some time and come back to low values when things are back to normal.

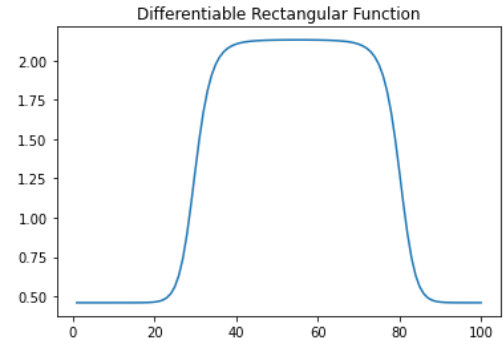

Fig. 4. A sample curve for a smooth continuous rectangle function, with time t (in days) plotted on X axis and the value of R0 plotted on Y axis. Here ai=30 and bi=50, which represents that after 50 days from ai, the curve falls.

The R0 value in the modified SEIR model is represented in (9)

$$R0 = R0_{Wave1} + R0_{Wave2} \qquad (9)$$

Here in (9) we consider the effect of two different waves of Covid-19 (such as for India). For additional waves, we can add the corresponding terms $R0_{Wave3}, R0_{Wave4}$ etc.

The R0 for each respective Covid-19 wave is given in (10):

$$R0_{Wavei} = \frac{R0_{start} - R0_{end}}{(1 + e^{-k*(-t+ai)*(-t+ai+bi)})} + R0_{end} \qquad (10)$$

Here, k defines the smoothness of the curve, a1 & b1 defines the start and end of the curve. R0-start and R0-end defines the max and min values of the curve.

$$R0 = \frac{R0_{start} - R0_{end}}{(1 + e^{k*(-t+a1)*(-t+a1+b1)})} + R0_{end} \qquad (11)$$

To model multiple waves of the infection, multiple such curves are used and it can be easily extended to any number of impacts.

The values of k, $a_i$, $b_i$, R0-start and R0-end are identified by fitting the function to real world Covid dataset.

R0 is designed to have multiple rectangle functions. In this work, 2 waves are considered, as is the case currently for India Covid-19 infections (first wave in 2020 and second wave in 2021). However, this can be easily extended to multiple waves. The individual rectangle function has parameters ai and bi, which controls the point of rise and fall.

When there is a lockdown, we assume that the spread (number of people who move from S to E within a day) reduces. So the R0 value, which is a function of time, is decreased during the lockdown. When we fit the equations to the actual data of covid infections during the lockdown, this will reflect automatically in the value of R0.

### B. SEIR model: Modified for effect of vaccination

Effect of vaccine is modeled by moving people from S to R directly. We assume people becomes immune after 30 days of receiving 1st dose of vaccine. 30 days interval is considered for body to build immunity against the virus. Weekly vaccine counts are obtained from CoWin public APIs [6], and the daily vaccine counts are obtained by dividing week count by 7.

Fig. 5 shows the plots of vaccination numbers for India, taken from [6].

In order to model the effect of vaccination on the population under consideration, we introduce an additional term V. Here, $V$ represents the population of the country that is vaccinated at a given time. Hence, V is a function of time. The modification is done in equation (1) mentioned earlier.

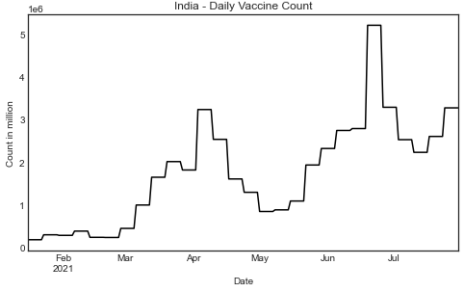

Fig. 5. Plot of the actual number of people vaccinated with at least one dose of Covid-19 vaccine, in India, varying with time (taken from Cowin app public APIs [6])

Terminology:

V Vaccination at time t

Rec Proportion of population recovering, including the effect of vaccine

$\frac{dRec}{dt}$ Population recovering as per enhanced SEIR model which **considers vaccination effect** V which represents the number of people who took the vaccine.

In our modified SEIR equation, $\frac{dRec}{dt}$ is given in (12):

$$\frac{dRec}{dt} = \gamma(1 - Prob_{ItoC}) * I + (1 - Prob_{CtoD})\ 1/6.5 * \min(Beds - C) + V \quad (12)$$

### C. SEIR model: Modified for effect of second infection

Some people who are infected with Covid-19 get 2nd infection after recovering from 1st infection or get infected after taking the vaccine. We have modified the SEIR model equations to accommodate these cases. In SEIR models, S monotonically decreases over time. In our modified equations for 2nd infections, S could slightly increase.

In order to model the effect of the second infection of Covid-19, meaning the population who had earlier recovered again after getting infected, we introduce a constant Rsus. Hence, the rate of 2nd infections is assumed to be a constant percentage Rsus of recovered, and is not fitted from data.

We keep the value of Rsus as 0.01 as an approximation, representing the assumption that approximately 1% of the people infected with Covid and recovered are subsequently re-infected.

$Rsus$ : A constant (between 0 and 1) representing the percentage of population that goes back to Suspectable compartment indicating chance of second infection, where the rest of the infected population recovers with permanent immunity.

The modified equations with Rsus are shown in (13) and (14):

$$\frac{dS}{dt} = \frac{-\beta ISV}{N} + Rsus * \frac{dRec}{dt} \quad (13)$$

$$\frac{dR}{dt} = (1 - Rsus) * \frac{dRec}{dt} \quad (14)$$

In equation (13), the updated $\frac{dR}{dt}$ represents the population that gains permanent immunity considering the effect of vaccine and 2nd infection together.

### IV. DATASET AND PARAMETERS USED

As mentioned earlier, we have taken the India Covid-19 data from John Hopkins dataset [5] and vaccination data from CoWin public APIs [6].

We fit our modified equations to the real world data for India.

Along with the parameters of R0 curve, few other parameters are also fitted with real world dataset.

Table 1 gives the values of the parameters used for the modified SEIR equations.

TABLE I. VALUES OF MODIFIED SEIR MODEL PARAMETERS

| Name of parameter | value | initial value | min | max |
| --- | --- | --- | --- | --- |
| R_0_start | 2.13470497 | 3 | 2 | 5 |
| k | 2.06340457 | 2.5 | 0.01 | 5 |
| a1 | 60.4533838 | 90 | 0 | 350 |
| b1 | 146.800862 | 90 | 0 | 350 |
| a2 | 252.861166 | 90 | 0 | 350 |
| b2 | 32.678741 | 90 | 0 | 350 |
| R_0_end | 0.45969314 | 0.9 | 0.3 | 3.5 |
| prob_I_to_C | 0.02359074 | 0.05 | 0.01 | 0.1 |
| prob_C_to_D | 0.21900041 | 0.5 | 0.05 | 0.8 |
| s | 0.00767634 | 0.003 | 1.00E-03 | 0.01 |

Table 2 gives the summary statistics of the curve fitting. We have fit the modified SEIR equations to actual Covid-19 data for India using lmfit [7]. Lmfit is a python library for least squares minimization for curve fitting.

TABLE II. STATISTICS OF CURVE FITTING

| Parameter | Value |
| --- | --- |
| fitting method | least_squares |
| # function evals | 196 |
| # data points | 585 |
| # variables | 10 |
| chi-square | 1.3049e+10 |
| reduced chi-square | 22694231.4 |

| Akaike info crit. | 9918.42202 |
|---|---|
| Bayesian info crit. | 9962.13814 |

## V. RESULTS OF THE CURVE FITTING

In this section, we provide the results of our curve fitting, using lmfit library [7], for the modified SEIR equation for the actual Covid data for India [5,6].

Fig. 6 shows the plot of the best fit curve compared to the actual data for Covid-19 mortalities in India. Figure 7 gives the plot of the model fitting.

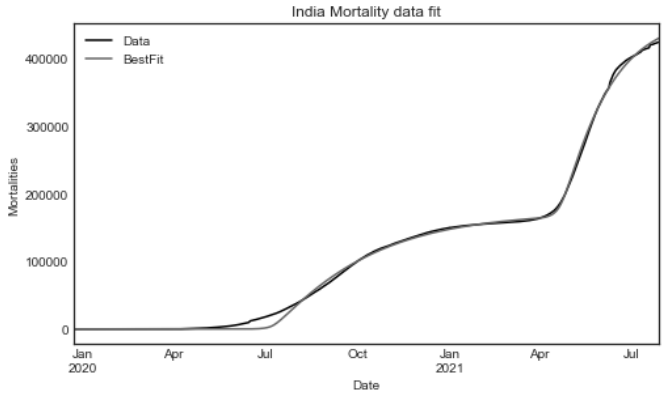

Fig. 6. Plot of the model fitting the SEIR model to actual India covid mortalities during the first and second wave

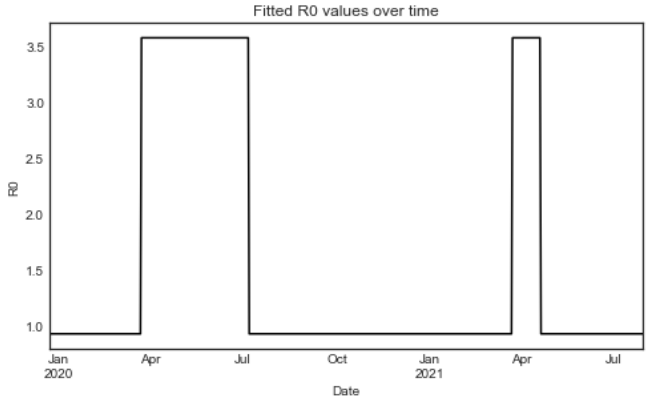

Fig. 7. Plot of the model fitting the SEIR model to actual India covid infections during the first and second wave. The first pulse represents the first wave in India in 2020 and the second pulse the second wave. The parameters a1, a2, b1, b2 represent the starting and width of the respective pulses.

Fig. 8 plots the values of S, E, I and R as a result of the curve fitting of the modified SEIR model to the actual Covid-19 data.

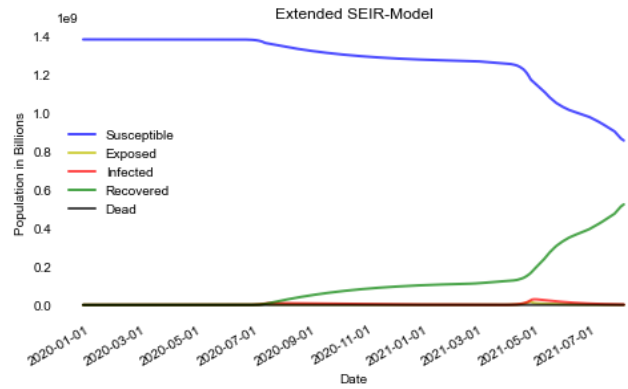

Fig. 8. Plot of the values of susceptible, exposed, infected, and recovered after fitting the modified SEIR model to the actual Covid-19 data for India.

Table 3 gives the results of the evaluation metrics such as Root Mean Square Error (RMSE), Mean Absolute Error (MAE), Mean Squared Error (MSE), R2 etc for measuring how well the modified SEIR model fits to the actual Covid-19 data for India after the curve fitting.

TABLE III. RESULTS OF THE EVALUATION METRICS FOR THE FIT OF ACTUAL COVID-19 DATA WITH THE MODIFIED SEIR MODEL

| Evaluation Metric | Value |
|---|---|
| R2 | 9.985000e-01 |
| Mean Absolute Error (MAE) | 2.978500e+03 |
| Mean Squared Error (MSE) | 2.230630e+07 |
| Root Mean Squared Error (RMSE) | 4.722954e+03 |
| Explained Variance | 9.986000e-01 |
| Max Error | 1.681597e+04 |
| Mean Squared Log Error | 5.665300e+00 |
| Median Absolute Error | 1.824912e+03 |
| Median Absolute Percentage Error | 4.386880e+02 |
| Akaike info crit. | 9.918422e+03 |

## VI. CONCLUSION AND FURTHER WORK

In this paper, we have proposed a few modifications of the SEIR model to accurately model the numbers for the Covid-19 pandemic for India, taking into account factors such as lockdown, vaccines and second infections. We have fitted our modified data with actual Covid-19 data for India, and obtained good results.

In future, we plan to extend our approach with different models for curve fitting and other machine learning models for regression. We also plan to implement the method for different states of India rather than taking India as a whole.